\documentclass[letterpaper, 10 pt, conference]{ieeeconf}  

\IEEEoverridecommandlockouts                              

\usepackage{graphics} 
\usepackage{epsfig} 
\usepackage{amsmath} 
\usepackage{amssymb}  
\usepackage{algorithm}
\usepackage{algorithmic}
\newtheorem{theorem}{Theorem}[section]

\newtheorem{lemma}[theorem]{Lemma}
\newtheorem{problem}{Problem}

\usepackage[noadjust]{cite}
\usepackage{todonotes}
\usepackage{hyperref}

\title{\LARGE \bf
Adaptive View Planning for Aerial 3D Reconstruction
}

\author{Cheng Peng
\thanks{
Cheng Peng and Volkan Isler are with the Department of Computer Science, University of Minnesota, Twin Cities, Minneapolis, MN, 55455}
\and
Volkan Isler
}

\begin{document}

\maketitle
\thispagestyle{empty}
\pagestyle{empty}

\begin{abstract}
With the proliferation of small aerial vehicles, acquiring close up imagery for high quality reconstruction is gaining importance. 
We present an adaptive view planning method to collect such images in an automated fashion. We first start by sampling a small set of views 
to build a coarse proxy to the scene. We then present (i)~a method that builds a set of adaptive viewing planes for efficient view selection
and (ii)~an algorithm to plan a trajectory that guarantees high reconstruction quality which does not deviate too much from the optimal one.
The vehicle then follows the trajectory to cover the scene, and the procedure is repeated until reconstruction quality converges or a desired level of quality is achieved.
The set of viewing planes provides an effective compromise between using the entire 3D free space and using a single view hemisphere to select the views.
We compare our algorithm to existing methods in three challenging scenes. Our algorithm generates views which produce the least reconstruction error comparing to three different baseline approaches. 
\end{abstract}

\section{Introduction}

3D scene reconstruction has been an active research topic for more than two decades.  The ability to obtain a rich and accurate 3D model is imperative for many applications including scene visualization, robot navigation, and image based rendering.  Recently, with the prevalence of small and nimble aerial vehicles (drones), scene reconstruction from close-up aerial imagery has been gaining popularity.  However,  the  battery  and  processing power of the drones limit image acquisition for large-scale, high-quality reconstruction.

The most common method to obtain aerial imagery in an automated fashion is to use an off-the-shelf flight planner, such as Pix4D~\cite{Pix4D}. These planners generate simple zig-zag or circular trajectories for coverage. As we will show, these can be insufficient to produce high-quality 3D reconstruction especially in the case of low altitude flights over complex scenes. On the other end of the spectrum, there are $\textit{next-best-view}$ approaches, reviewed in the next section, which choose views actively so as to increase the amount of acquired information~\cite{nbv_pe,dunn2009next}. These approaches are hard to implement on existing systems due to the need for significant on board processing to accurately localize on the go with a real-time SLAM approach~\cite{mur2015orb,klein2007parallel,engel2014lsd}, a decision making mechanism running on board to decide on the next best view, and a controller to execute the strategy.

\begin{figure}
    \centering
    \includegraphics[width=0.5\textwidth]{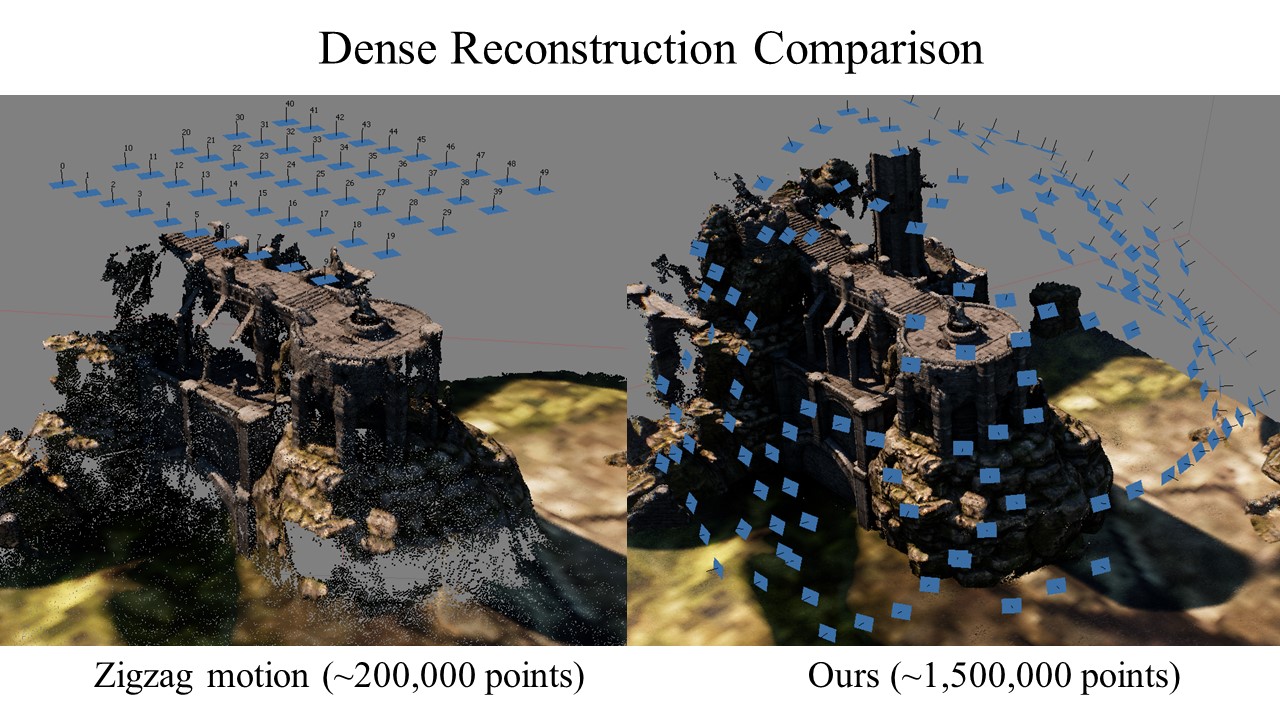}
    \caption{Dense point cloud comparison between the baseline zig-zag motion and our method ($2^{nd}$ visit). Note that our method produces significantly denser point clouds comparing to that of the baseline.}
    \label{fig:dense_comp}
\end{figure}

As a result, there has been recent interests in $\textit{explore-then-exploit}$ methods~\cite{roberts2017submodular, hepp2017plan3d} for multi-stage image acquisition. These methods first ``explore" the environment using a fixed trajectory (e.g. a zig-zag motion as shown on the left of Fig~\ref{fig:dense_comp}). The images are then used to build a rough mesh or voxel representation which is then used to plan an ``exploit" trajectory that maximizes the coverage and reconstruction quality of the scene in a second pass. Both trajectories are executed in an open loop fashion without feedback. The advantage of this approach is that the secondary exploit visit to the scene has prior knowledge of the general geometry,  making it possible to optimize for coverage and accuracy globally.  Yet, it is easy to execute because the view points are generated offline and visited using a standard way-point navigation algorithm.

In this paper, we propose a novel trajectory planning algorithm that naturally adapts to scene geometries and produces high quality reconstructions. Our novel adaptive viewing planes (rectangles) also provides an effective alternative to selections in 3D free space~\cite{roberts2017submodular, hepp2017plan3d, hoppe2012photogrammetric}. 
Our results show that, in contrast to existing ``explore and exploit" methods which collect only two sets of views, 
reconstruction quality can be drastically improved by adding a third set. 
They also indicate that three rounds of data collection is sufficient even for very complex scenes.

One of the key contribution of our paper is that we formulate the trajectory planning problem into a new variant of Traveling Salesperson Problem. We also provide a constant factor theoretical bound to our trajectory length comparing to the optimal one.

\begin{figure*}
\centering
	\includegraphics[width=0.8\textwidth]{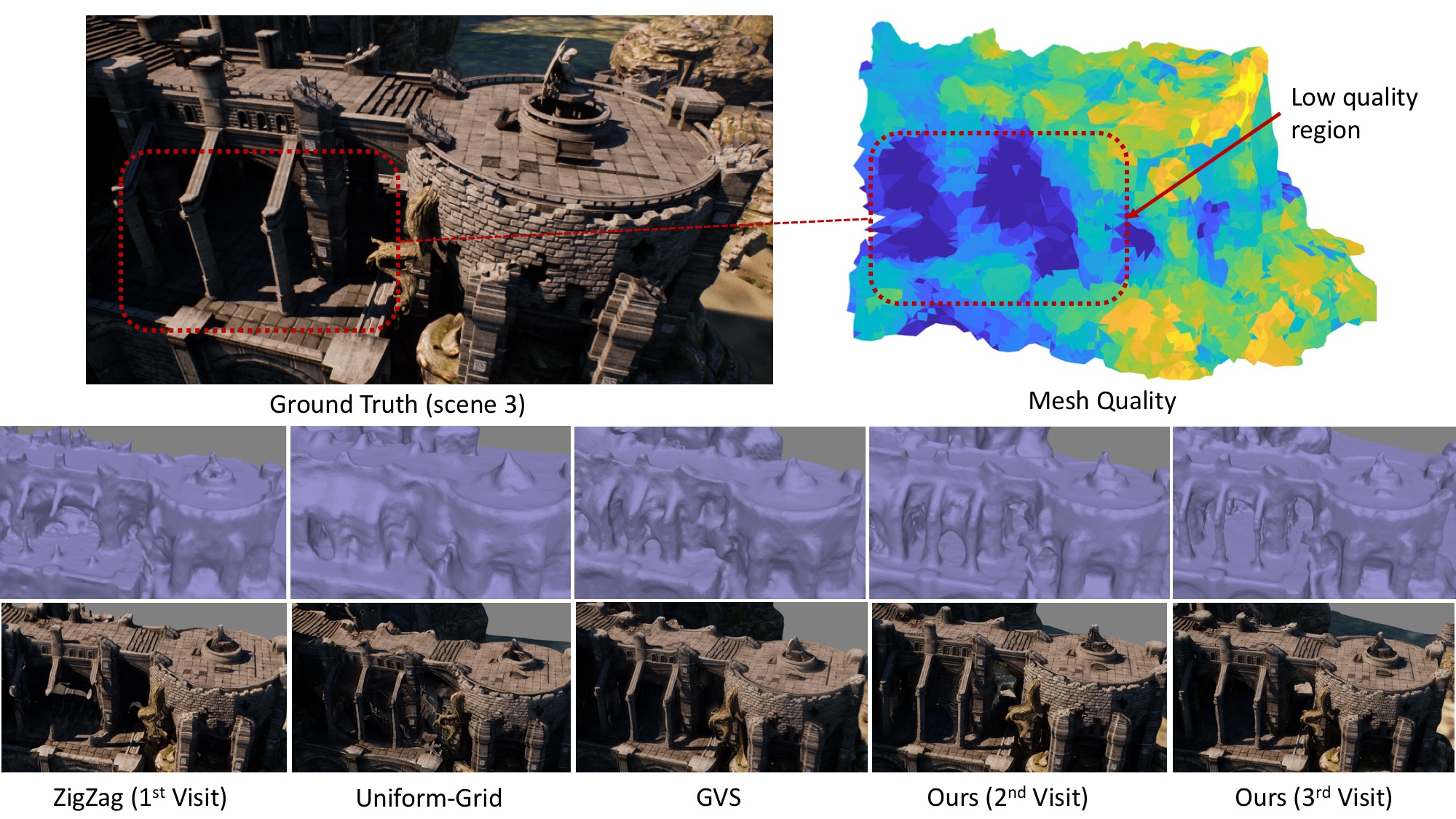}
	\caption{The effect of view plans on reconstruction quality. The top left figure shows the scene. Bottom images provide a qualitative comparison of  zigzag motion, uniform-grid method, greedy view selection, and our method after $2^{nd}$ and $3^{rd}$ visits. Rendered mesh and texture map views are shown to provide both structural and visual comparisons. Our method visually outperforms all other methods. The $3^{rd}$ visit improves the reconstruction of low quality regions (top right) significantly.}
	\label{fig:lqvCompare}
\end{figure*}

\section{Related Work}
%
The view selection problem remains an essential topic for 3D reconstruction. One of the early view selection methods optimizes the baselines among a set of images for accurate depth reconstruction~\cite{farid1994view}. Maver and Bajcsy~\cite{maver1993occlusions} utilize the knowledge of contours and occlusions to choose views. 
Scott et al.~\cite{scott2003view} later analyze the view selection problem for reconstruction in detail and show that it is an NP-Complete problem. 
This line of work is called active vision because the views are managed actively to improve reconstruction quality~\cite{chen2008active}. 
Scott et al.~\cite{scott2003view} utilize an integer programming method to solve for the view planning problem. Vasquez et. al.~\cite{nbv_pe} propose a greedy selection method that optimizes for the next-best-view for reconstruction. 

To reduce view search space, the view sphere method~\cite{vasquez2009view} is proposed to limit the view selection around an ``enclosing" sphere to an object. However, a single view sphere cannot handle large scaled scenes. 
Therefore, other methods~\cite{roberts2017submodular,hepp2017plan3d} discretize the 3D viewing space where the number of potential views in the free space can be too large to compute efficiently~\cite{roberts2017submodular}.
Instead of choosing the views in 3D space, we propose a set of adaptive viewing rectangles that reduces the search space from 3D to 2D, which also reduces the number of potential views. 

To address battery limitation of a drone, Robert et. al.~\cite{roberts2017submodular} constrain the traveling cost during their reward collection process but provide no theoretical guarantees. Tokekar et. al.~\cite{iram2012coverage} formulate the problem of coverage into a Traveling Salesperson Problem (TSP), which is NP-hard~\cite{arora1997nearly}. The advantage is that theoretical guarantees can be provided for the trajectory length. Their problem setup is on a 2D plane where they provide a three approximation ratio. For visiting regions in 3D, Plonski and Isler~\cite{plonski2016approximation} formulate the coverage of a drone as a Cone-TSPN (TSP with Neighborhoods of cones) problem. However, their algorithm also applies for planar regions that guarantees coverage which is not suitable for us. Since triangulating a point requires more than a single view, the TSPN formulation is not sufficient to solve our problem. 
Therefore, the novelty of our contribution is that we reformulate the view planning problem as a $2.5D$ TSP on the adaptive viewing rectangles. We also provide a constant theoretical bound ($\leq 18$) to the trajectory length while guaranteeing high quality reconstruction. 

%


\section{Problem Formulation}
The objective is to find a minimum length trajectory for a drone to take images of an unknown scene such that the final 3D reconstruction model is of high quality. 

Since the scene is unknown, we assume a geometric proxy is initially given as a triangular mesh $\mathcal{M = \{\mathcal{F},\mathcal{V}\}}$ built on a point cloud $\mathcal{I} = \{I_1,I_2,...,I_{\mathcal{I}}\}$ with mesh vertices $\mathcal{V} = \{v_1,v_2,...v_\mathcal{V}\}$ and a set of faces $\mathcal{F} = \{f_1,f_2,...,f_\mathcal{F}\}$, $f_i \in v \times v \times v$. 
We would like to find a feasible trajectory $\mathcal{J} = \{\{s_1,\phi_1\},\{s_2,\phi_2\},...,\{s_n,\phi_n\}\}$ with positions $S = \{s_1,s_2,...,s_n\}$ and orientations $O = \{\phi_1,\phi_2,...,\phi_n\}$ that covers $\mathcal{F}$ with high quality, where $s \in \mathbb{R}^3$ and $\phi \in \mathbb{R}^3$.
We define the quality measurement as $Q(f, \mathcal{J})$ (Section~\ref{sec:lqr}) for a set of views in $\mathcal{J}$ and a face $f \in \mathcal{F}$. We also define the visibility between a face $f$ and a view $s$ as $Vis(f,s)$ (Section~\ref{sec:lqr}) where $Vis(f,s)=1$ indicates that $f$ is visible from $s$ and $0$ otherwise.
Denote $N(f)$ and $C(f)$ as the normal vector and center location of the face $f$ respectively. 
Here we define the problem formally. 
\begin{problem}\label{prob:formulation}
Given a triangular mesh $\mathcal{M = \{\mathcal{F},\mathcal{V}\}}$ as a geometric proxy of the scene, we would like to obtain a trajectory $\mathcal{J} = \{\{s_1,\phi_1\},\{s_2,\phi_2\},...,\{s_n,\phi_n\}\}$ with positions $S = \{s_1,s_2,...,s_n\}$ and orientations $O = \{\phi_1,\phi_2,...,\phi_n\}$ such that 
\begin{equation}\label{eq:problemform}
\begin{split}
    \min(len(\mathcal{J})) \\
    s.t. \ |\kappa(f,\mathcal{J})| \geq t&, \forall{f} \in \mathcal{F} \\
    Q(f,\mathcal{J}) \geq Q^*&, \forall{f} \in \mathcal{F}\\
    |\mathcal{J}| \leq B& \\
    \end{split}
\end{equation}
where $len(\mathcal{J})$ is the trajectory length of $\mathcal{J}$. $\kappa(f) = \{s | Vis(f,s) = 1, \forall s \in \mathcal{J}\}$ indicates a set of views that are visible to face $f$ and $t \geq 2$.
$Q^*$ is the predefined quality threshold that limit the viewing angles to $\theta \in [\theta_{min}, \theta_{max}]$, which is defined in Eq~\ref{eq:quality}.
We also limit the number of views in the trajectory to be less than $B$ to account for computational cost of Structure from Motion.
We assume the angle of the field of view for our camera is $\pi/2$.
\end{problem}

\section{Reconstruction Quality}\label{sec:lqr}
The quality of the final dense reconstruction depends on various factors including occlusion boundaries, triangulation angles, the number of correctly matched image features, and the relative image resolutions. In this section, we address those factors and formulate them into our problem formulation in Eq~\ref{eq:problemform}. 

\subsubsection{Triangulation angle}
Generally speaking, large parallax to a point results in accurate triangulation. In fact, Peng and Isler~\cite{peng2017recon} proved that viewing the target at $90^\circ$ can minimize the triangulation uncertainty. However, large viewing angle reduces matching accuracy.
Naturally, when the views are nearby (small parallax), the image patches (features) tend to be similar, which can increase matching accuracy. However, small parallax induces large depth uncertainty since the triangulated points can move along the viewing ray without changing the pixel location too much~\cite{peng2017recon,schonberger2016pixelwise}. 

To address this issue, we use the heuristic in~\cite{roberts2017submodular} and~\cite{furukawa2015multi} which bound the triangulation angle to be $\theta \in [5^\circ, 20^\circ]$.
In addition to that, we also propose a quality measurement $Q(f,\mathcal{J}) = \frac{\sin(\theta)}{d_1d_2}$ for face $f$ as the inverse of the uncertainty measurement based on Bayram et. al.~\cite{bayram2016gathering}, where $d_1,d_2$ are the distance from two views to the triangulated points with triangulation angle $\theta$. For a set of views $\kappa(f)$ that sees the target $C(f)$ with maximum angle $\Theta(f) = \max_{s_i,s_j \in \kappa(f)} \angle(\overline{s_iC(f)},\overline{s_jC(f)})$ and $\{s_i,s_j\} = \arg\max\Theta(f)$, the quality is defined as 
\begin{equation}\label{eq:quality}
    Q(f,\mathcal{J}) = \sin{\Theta(f)}/(\overline{s_iC(f)} \cdot \overline{s_jC(f)}) \geq Q^*
\end{equation}

\subsubsection{Image resolution}
For views with significant distance variations to the target, the difference between resolutions can pose difficulties to both sparse and dense reconstruction~\cite{zheng2014patchmatch,schonberger2016pixelwise}. For sparse reconstruction, one can build an image pyramid~\cite{adelson1984pyramid} to account for resolution variations. For dense reconstruction, similar image resolutions and patch shapes can increase the similarity score between the matching pieces~\cite{zheng2014patchmatch, schonberger2016pixelwise}. 
 
To favor consistent resolutions of the views, we integrate a viewing distance to be within $[d-\epsilon_d,d+\epsilon_d]$ with some tolerance $\epsilon_d$. We first define the visibility of a face from a view $s$ as $Vis(f,s)$. A view $s \in \mathcal{J}$ is considered visible to a face $f$ if two conditions are satisfied. The first condition is that $s$ should be in line-of-sight with $C(f)$ and the second one is that $\overline{sC(f)} \in [d-\epsilon_d,d+\epsilon_d]$. Therefore, $Vis(f,s)=1$ when both conditions are satisfied and 0 otherwise.
We assume the field of view of our camera is $\pi/2$ and set $\epsilon_d \leq (\sqrt{2} - 1)d/2$ to account for features at the image boundary that are of maximum distance to the camera center.

\subsubsection{Feature correctness}
It is not sufficient to maintain a high quality reconstruction with only the quality constraint $Q^*$. The extracted image features can still be wrongly associated. To avoid triangulating the wrong matches, the standard method is to perform a geometric consistency check~\cite{furukawa2015multi,engel2014lsd,klein2007parallel,mur2015orb} with RANSAC~\cite{fischler1981random}. However, if the scene is a plane, this method will fail due to the matching ambiguities along the epipolar line~\cite{luong1996fundamental}.

Therefore, more than two views are desired for robust triangulation. 
During the feature matching process, additional detected features are used as a validation process to ensure the detection accuracy.
To formalize this, we define $\kappa(f,\mathcal{J}) = \{s | Vis(f,s) = 1, \forall s \in \mathcal{J}\}$ as a set of visible views for the face $f$ from a given trajectory $\mathcal{J}$.
\begin{equation}\label{eq:kappa}
    |\kappa(f,\mathcal{J})| \geq t, \forall f \in \mathcal{F}
\end{equation}
where $t \geq 2$ is a constant.  




\section{Trajectory planning}
To obtain high-quality reconstruction, the trajectory planning algorithm needs to satisfy both Eq~\ref{eq:quality} and Eq~\ref{eq:kappa}.
If we set $|\kappa(f,\mathcal{J})| = 2$ and ignore the triangulation angle constraint, then only two views are required for each face. This trajectory planning problem can be viewed as a set cover problem that finds the minimum number of views such that their common region covers the entire scene, which is shown to be NP-Complete~\cite{tarbox1995planning}. The problem is further complexed by optimizing over both position and orientation for each view.  
In real applications, more than 2 views are desired to eliminate the matching error per region, which adds another layer of difficulty. 

To address the view planning problem, we first introduce the adaptive viewing rectangles for efficient view selection. Then we plan a back-and-forward sweeping motion for each rectangle and prove that the resulting trajectory length deviates only a constant factor compared to the optimal one. 

\subsection{Adaptive Viewing Rectangles}\label{sec:avp}
\begin{figure}
    \centering
    \includegraphics[width = 0.35\textwidth]{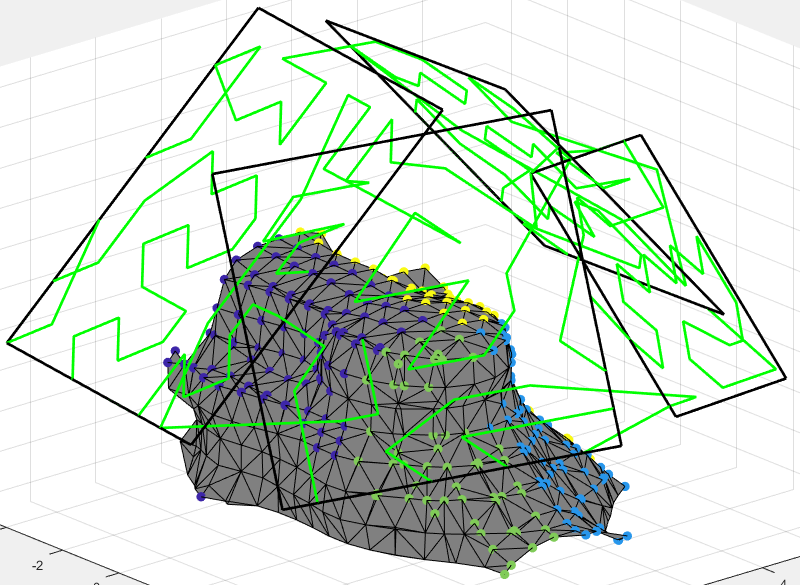}
    \caption{The adaptive viewing rectangles for different clusters of the scene patch in different color and the final trajectory computed through the rectangular grid.}
    \label{fig:tsp_traj}
\end{figure}


We propose a set of viewing planes that adapts to the scene geometry, for which we called the Adaptive Viewing Rectangles (AVR) as $\mathcal{P}= \{P_1, P_2,...,P_k\}$ (Fig~\ref{fig:tsp_traj}). For each viewing rectangle $P_i \in \mathcal{P}$, the views are located relatively parallel to the scene so that the viewing angle and image resolution to the same patch are relatively consistent. Comparing to views selected in 3D free space, our AVR eliminates the necessity to compute viewing distance for each view and provides less potential views for selection. 

More specifically, we first cluster the faces $\mathcal{F}$ using k-mean clustering method~\cite{teknomo2006k} based on the faces' locations denoted as $\mathcal{C} = \{C_1,C_2,...,C_k\} \subseteq \mathcal{F}$, where $C_i$ is a cluster of faces. 
Next, we construct the adaptive viewing rectangles $\mathcal{P} = \{P_1,P_2,...,P_k\}$ for the patch clusters $\mathcal{C}$. 
More specifically, for each elevated cluster $C_i + d\cdot\mu_N(C_i)$, we find a fitting plane $\hat{P_i}$ using a standard least square fitting method, where $\mu_N(C_i)$ is the mean normal direction of all points in $C_i$. The rectangle $P_i$ is the minimum fitting rectangle of the projections of $C_i$ to $\hat{P_i}$ as shown in Fig~\ref{fig:tsp_traj}. 

\subsection{Viewing grid method}
Many methods~\cite{roberts2017submodular,hepp2017plan3d,peng2017recon} define the visibility cones (hemisphere) such that the target is visible from views inside of the cones. However, it is unclear how to define the angle or bisector direction of the cones for each target. Therefore, instead of modeling specifically the visibility of each target as a geometric primitive, we impose a viewing grid on each viewing rectangle $P_i$ such that each grid point should be visited. 

Similar to the viewing grid method for reconstruction introduced by Peng and Isler~\cite{peng2017recon}, the rectangular viewing grid $G_i$ is a collection of view points on rectangle $P_i$ with resolution $r$. We assume that the widths of any rectangles in $\mathcal{P}$ are larger than resolution $r$.
From Theorem 5.3 in~\cite{peng2017recon}, the imposed viewing grid with resolution $r=d$ only requires less than $Area(P_i)/d^2$ of views and achieves low ($\leq 1.72$ comparing to optimal) reconstruction uncertainty for the scene. The view orientation $\phi$ is set to the direction of the normal of the rectangle.

\section{Algorithm}
Given the adaptive viewing rectangles $\mathcal{P}$ for the patch clusters $\mathcal{C}$, we impose a viewing grid $G_i$ for each rectangle $P_i$ as a collection of sparse view points with resolution $r = d$. Denote $\mathcal{G} = \{G_1,G_2,...,G_k\}$ as a collection of all the viewing grids. 
\subsection{Algorithm outline} \label{sec:alg}
The corresponding TSP tour is computed as follows.
\begin{itemize}\label{alg:tour}
    \item Compute a back-and-forth tour that traverses the viewing grid points with resolution $r$ in $G_i$ and denote the tour as $T_i$ for the rectangle $P_i$ .
    \item Compute the minimum distance between any two distinct viewing grids $\{G_i,G_j\}$ with the distance matrix $D$ and compute a minimum spanning tree $MST(\mathcal{G})$ from $D$ where $e_{ij} \in MST(\mathcal{G})$ indicates the edge that connects the viewing grids $\{G_i,G_j\}$. 
    \item Given each tour $T_i$ on the rectangle and duplicated each edge in the minimum spanning tree $MST(\mathcal{G})$, compute a Euclidean tour that visits all $T_i$ and edges denote the final tour as $T_f$.
\end{itemize}
To improve the reconstruction quality, we refine the 3D model by iteratively identifying the low quality regions and plan trajectory that only visits those regions. 

\subsection{Algorithm analysis}
In this section, we will analyze the performance of our proposed algorithm and show that the worst case performance does not deviate too much from the optimal solution. The optimal trajectory is $T^*$ and its length $l^*$ is the shortest trajectory that satisfies the constraints in Eq~\ref{eq:problemform}. 
Then, we can claim the following for our trajectory.
\begin{theorem}\label{thrm:traj}
    Given the trajectory $T_f$ constructed in Section~\ref{sec:alg} with length $l_f$, the optimal trajectory $T^*$ with length $l^*$, and the viewing angle $\theta \in [\theta_{min}, \theta_{max}]$, 
    \begin{equation}
        l_f \leq (12\frac{d}{r} + 6) l^*
    \end{equation}
\end{theorem}

The approximation ratio is $\leq 18 l^*$ given the resolution of our approach is set to $r=d$,
To prove the theorem, we first bound the optimal trajectory $l^*$ with respect to the areas of the rectangles in $\mathcal{P}$. 
\begin{lemma}
Denote the area of each rectangle $P_i$ as $Area(P_i)$. Given the viewing distance to the scene to be $d$, 
\begin{equation}\label{eq:area}
    \sum_{P_i \in \mathcal{P}}Area(P_i) \leq 4d l^*
\end{equation}
\end{lemma}
This Lemma can be proved using the packing theorem~\cite{dumitrescu2003approximation} and the details are shown in~\cite{icra2019traj}

\begin{lemma}
Given a minimum spanning tree $MST(\mathcal{P})$ where $e_{ij} \in MST(\mathcal{P})$ indicates the edge that connects the rectangles $\{P_i,P_j\}$, we claim that
\begin{equation}
    \sum_{e \in  MST(\mathcal{P})}|e| \leq l^*
\end{equation}
\end{lemma}
The intuition is that the optimal trajectory has to visit each plane $P_i$ at least once in order to cover the scene with desired distance. Therefore, the shortest distance ($\sum_{e \in  MST(\mathcal{P})}|e|$) linking each rectangle becomes a lower bound for the optimal trajectory. 

After providing the lower bound for the optimal trajectory, we now analyze each component in our trajectory $T_f$.
It is easy to see that our trajectory $T_i$ with length $l_i$ covers at most three sides of the grid cell of resolution $r$. 
\begin{equation}
    l_i \leq 3r \frac{ Area(P_i)}{r^2}
\end{equation}

We can then write the final trajectory length $l_f$ as follows:
\begin{equation}
    \begin{split}
        l_f \leq& \sum_{P_i \in \mathcal{P}}Area(P_i) \frac{3r}{r^2} + 2\sum_{e \in MST(\mathcal{G})}|e| \\
        \leq& (12\frac{d}{r} + 6)l^*
    \end{split}
\end{equation}


For more detailed version of the proofs, please refer to the technical report~\cite{icra2019traj}.
\begin{figure*}
\centering
	\includegraphics[width=0.8\textwidth]{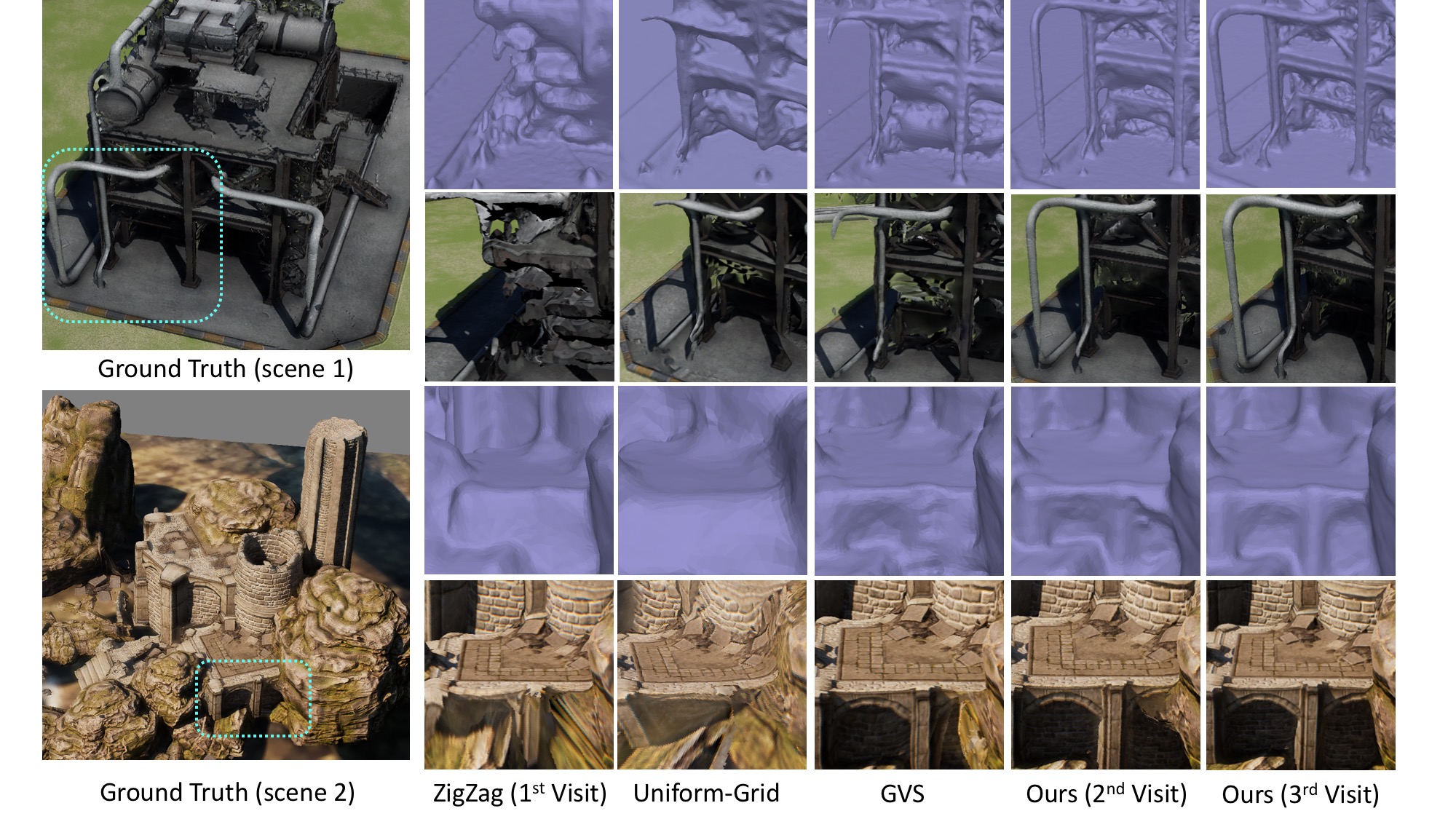}
	\caption{Qualitative comparison of the reconstruction among zigzag motion, uniform-grid method, greedy view selection, and our method for the $2^{nd}$ and $3^{rd}$ visits. Both the rendered mesh and texture map are shown to provide both structural and visual comparison. Our method out performs all other methods, where the $3^{rd}$ visit improves the structure further more.}
	\label{fig:closeupCompare}
\end{figure*}
\section{Results}
\begin{table*}[t]
\centering
\begin{tabular}{|c||c|c|c||}
\hline
 Methods & Depth Error Avg(mm) & Depth Error Std(mm) & Completeness($\%$)\\
\hline
ZigZag & 211.5 & 203.4 & $14.5\%$\\ 
\hline
Uniform-Grid & 189.3 & 143.5 & $19.3\%$\\ 
\hline
GVS & 143.9 & 120.3 & $18.4\%$\\ 
\hline
$\textbf{ours}$($2^{nd} visit$)& $\textbf{114.3}$ & $\textbf{105.7}$ & $\textbf{29.5\%}$\\ 
\hline
$\textbf{ours}$($3^{rd} visits$)& $\textbf{104.9}$ & $\textbf{103.2}$ & $\textbf{29.8\%}$\\ 
\hline
\end{tabular}
\caption{The comparison of depth error and completeness}
\label{tab:depthComp}
\end{table*}

In order to evaluate our method, we use a high-quality visual rendering software, Unreal Engine~\cite{unrealengine}. It is a game development engine that produces photo-realistic scenes. We conduct our experiment in synthetic environments by controlling a virtual camera using the UnrealCV Python Library~\cite{qiu2017unrealcv}.
We tested our method of view planning in 3 different synthetic scenes (from GRASS LAND~\cite{grassland} and OIL Refinery~\cite{oilrefinery} data sets) that contain occluded regions from the top view. 
We compare our reconstruction results both qualitatively and quantitatively among the results of three baseline methods.
Qualitatively, we show both the mesh and the texture map to demonstrate the visual completeness of the reconstruction. Quantitatively, we perform both accuracy and coverage test using the dense reconstruction results with the ground truth depth data.

\subsection{Implementation Details}

We perform dense reconstruction, mesh generation and texture mapping using a commercial software Agisoft~\cite{agisoft}. To produce the depth reconstruction for each image, we use the state-of-the-art algorithm COLMAP~\cite{schoenberger2016mvs}. For each scene, the initial zigzag coverage spacing is set to 1 meters and the distance to the ground is set to 20 meters. The views along the trajectory are selected based on the coarse-to-fine selection method proposed by Peng and Isler~\cite{peng2017recon}. The quality threshold is set to $Q^* = \frac{\sin(45\circ)}{(\sqrt(2)5)^2} = 0.014$ and the computational cost of the views are set to $B = 300$, which is sufficient for the trajectory to generate high quality reconstructions. In cases when the two rectangles are intersecting, then the line of the intersection will separate the two rectangles into four pieces. We merge the intersecting rectangles by taking the largest piece in each rectangle instead. 
In our experiments, the views selected from the final trajectories ranges from 100-250 images depending on the size of the scene. 

\subsection{Comparison Methods}
To compare our view planning algorithm, we build three different baseline methods. There is no benchmark data-set that allows active view selection for large scale aerial 3D reconstruction. Therefore, we generalize the core idea from other methods~\cite{roberts2017submodular,hepp2017plan3d} and implement our representative versions.  
\subsubsection{ZigZag}
The basic trajectory to cover a scene is a zigzag motion. We plan this motion with a predetermined height that fully covers the scene. The zigzag motion is available using a commercial flight planner such as Pix4D~\cite{Pix4D}. 
\subsubsection{Uniform-Grid}
Aside from the basic zigzag motion, another naive baseline method is to reconstruct a uniformly spaced sparse views in 3D viewing space. We construct this Uniform-Grid method in a discretized 3D viewing space. The number of the sparse views will be equal to that of our method for comparison while the resolution is set to 1 meter.
\subsubsection{Greedy View Selection}
Many different methods~\cite{roberts2017submodular,hepp2017plan3d} model the cost function in view planning problem as a submodular function, where the views are selected using the greedy algorithm. 
Therefore, we implement a naive Greedy View Selection (GVS) method to compare with our method. 
For each view $s$ with orientation $\phi$, we define the gain of the view as the total region quality. 
\begin{equation}\label{eq:gain}
	Gain(c) = \sum_{f \in \chi(s)} Q(f)
\end{equation}
Given a set of views $C \subseteq S$, the marginal gain of an additional view $s$ is the following. 
\begin{equation}\label{eq:gvs}
	Gain(s, C) = \sum_{f \in \chi(C \bigcup s)-\chi(s)}Q(f)
\end{equation}
where $C \subseteq S$ is a subset of views in $S$ and $\chi(C)$ are the faces covered by the view set $C$.

The GVS method has access to our AVR and selects the next best view using Eq~\ref{eq:gvs} from its neighbor views. We define the neighbor views to be within 1 meter. The GVS method starts from a random view point and terminates when the total number of views exceeds that of our algorithm.

\subsection{Qualitative Comparison}
As shown in Fig~\ref{fig:lqvCompare} and Fig~\ref{fig:closeupCompare}, we can see that our method outperforms all other methods. In Fig~\ref{fig:lqvCompare}, the third visit to the same scene shows significant improvements of the low quality regions. Those improved regions are not reconstructed from the initial scene and they are only identified through the second visit to the scene. 
It validates our assumption that the initial estimated geometry is too coarse for view planning. A closeup comparison is shown in Fig~\ref{fig:closeupCompare}. It is evident that our method produces significantly better structure. 

\begin{figure}
\centering
	\includegraphics[width=0.5\textwidth]{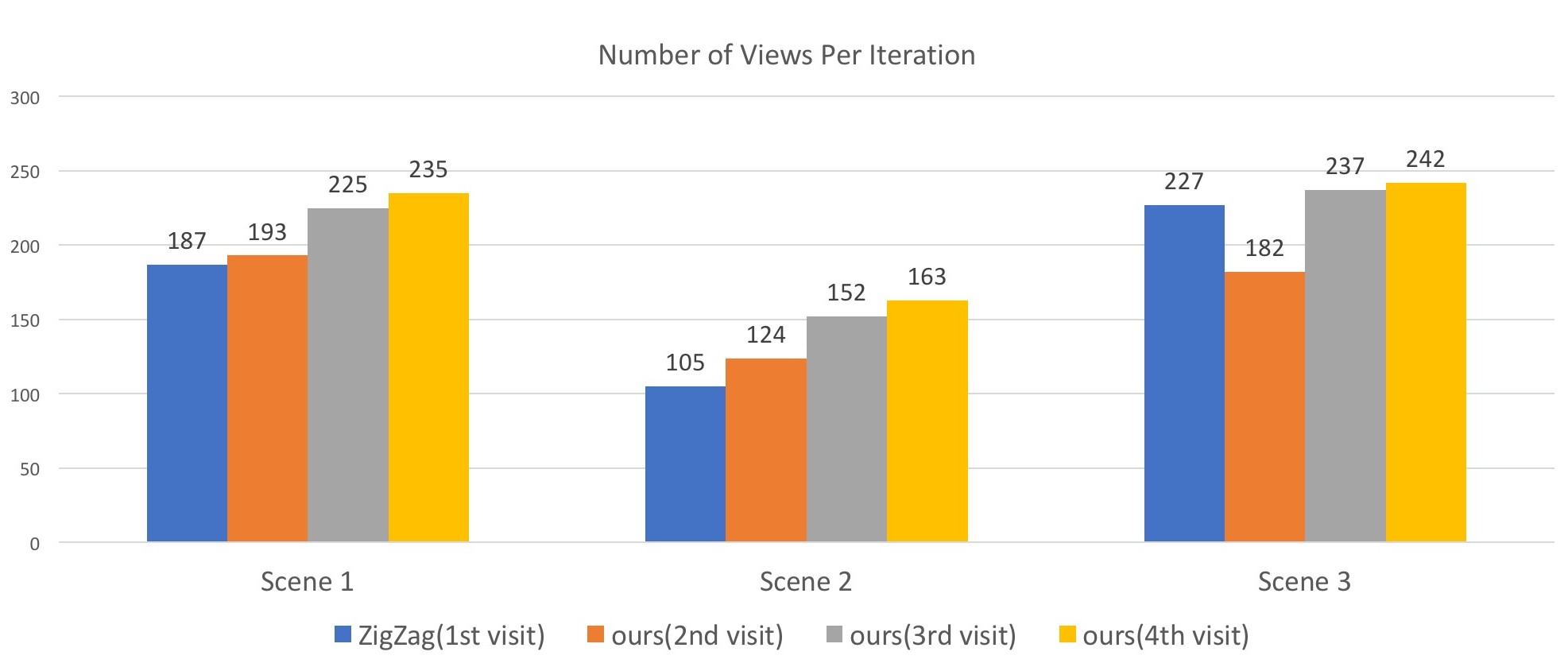}
	\caption{The number of input images at each iteration: the first iteration is the standard zigzag motion. The small increment in the number of views means that the low quality regions are well covered after the $2^{nd}$ and $3^{rd}$ iteration}
	\label{fig:iternum}
\end{figure}

We also show that the number of views for each iterative visit as in Fig~\ref{fig:iternum}. The first iteration is the standard zigzag motion. We can see that the $3^{rd}$ and $4^{th}$ visits only add less than 40 views to the $2^{nd}$ visit. The small increment in the number of views means the low quality regions are well covered after the $2^{nd}$ and $3^{rd}$ iteration. If the low quality regions appear far apart, the trajectory will contain only views for the desired regions. The input number of views is only dependent on the area of the low quality region.

\subsection{Quantitative Comparison}
We compare the reconstruction results quantitatively using the absolute depth error since the true depth can be obtained from UnrealCV Python Library~\cite{qiu2017unrealcv} in the simulation environment. 
The quantitative comparison of the reconstruction are shown in Table~\ref{tab:depthComp}.
We estimate the accuracy and coverage of the scene using the depth accuracy and dense point clouds completeness defined here. 
For accuracy, we calculate mean and standard deviation of the depth on the corresponding pixel from estimated dense point clouds and the ground truth depth image. 
The completeness is defined as the ratio between the number of pixels that contains the depth from dense point clouds and that of the ground truth. 
To analyze the depth from our scene, we filter out pixels that are more than 50 meters away in the ground truth. Since we are selecting only the medium-quality for our dense point clouds, the resulting completeness percentage can only be as high as $35\%$.
 
As shown in Table~\ref{tab:depthComp}, our method achieves the lowest depth error with the highest completeness percentage. Note that the $3^{rd}$ visit only increases the completeness of the $2^{nd}$ visit by less than $1\%$. This is because most of the scene is already covered and the additional iteration only increases the corresponding accuracy of the scene. 
\section{Conclusion}
This paper studies the problem of view selection for aerial 3D reconstruction. 
We propose an adaptive view planning method to reconstruct a scene from aerial imagery. 
Our work has three novel aspects: First, we present a method to reduce the view search space to a 2.5D viewing rectangles. Our adaptive viewing rectangles generalize the idea of using view sphere for a single object to more complex scenes. Second, we identify low reconstruction quality regions to plan our next set of views iteratively. We observe that the initial scene proxy reconstructed from the standard zigzag motion can be insufficient for planning. Therefore, more than 2 iterations of view planning are sometimes necessary to build a high quality reconstruction. At last, we show that the trajectory length of our proposed algorithm only deviates from the optimal one by a constant factor as low as $18$. 
We compare our view planning method with 3 baseline methods  using ground truth obtained from a photo-realistic rending software. Our method outperforms the baseline methods both qualitatively and quantitatively. 
\section*{Acknowledgement}
The author of this paper would like to thank Wenbo Dong and Shan Su for their help on the algorithm and setting up the virtual simulation. This work was supported by NSF Awards 1525045 and a Minnesota State LCCMR grant.

\addtolength{\textheight}{-7cm}   

\bibliographystyle{splncs}
\bibliography{egbib}


\end{document}